\documentclass[lettersize,journal]{IEEEtran}
\usepackage{amsmath, amsthm,amssymb,subfigure,cite,color,booktabs,topcapt,times,multirow}
\usepackage[T1]{fontenc}
\usepackage{enumerate}
\usepackage{cite}
\usepackage{url}
\usepackage{algpseudocode}
\usepackage{algorithmicx}
\usepackage{gensymb}
\usepackage[linesnumbered,ruled,vlined]{algorithm2e}
\usepackage{amssymb}
\usepackage{courier}
\usepackage{soul}
\usepackage{amsfonts,amssymb}
\usepackage{lineno,amsmath}

\usepackage{booktabs}
\usepackage{multirow}
\usepackage{pifont}
\usepackage{makecell}

\usepackage{array}
\usepackage[caption=false,font=normalsize,labelfont=sf,textfont=sf]{subfig}
\usepackage{textcomp}
\usepackage{stfloats}
\usepackage{verbatim}
\usepackage{graphicx}
\begin{document}
\bstctlcite{myBSTcontrol}
\title{Air-Ground Collaborative Robots for Fire and Rescue Missions: A Survey from the Mapping and Navigation Perspective}

\author{\IEEEauthorblockN{Ying~Zhang,~\IEEEmembership{Senior Member,~IEEE,}
                          Haibao Yan, Danni Zhu, Jiankun Wang,~\IEEEmembership{Senior Member,~IEEE,} \\
                          Cui-Hua Zhang, Weili Ding, Xi Luo, Changchun Hua,~\IEEEmembership{Fellow,~IEEE}, and Max Q.-H. Meng,~\IEEEmembership{Fellow,~IEEE}  
                          }

\thanks{Y. Zhang, H. Yan, D. Zhu, C. Zhang, W. Ding, X. Luo, and C. Hua are with the School of Electrical Engineering, and the Hebei Key Laboratory of Intelligent Rehabilitation and Neuromodulation, Yanshan University, Qinhuangdao, 066004, China. (e-mail: yzhang@ysu.edu.cn; hby97@stumail.ysu.edu.cn; zdn@stumail.ysu.edu.cn; cuihuazhang@ysu.edu.cn; weiye51@ysu.edu.cn; luox@ysu.edu.cn; cch@ysu.edu.cn).

J. Wang, and Max Q.-H. Meng are with Shenzhen Key Laboratory of Robotics Perception and Intelligence, and the Department of Electronic and Electrical Engineering, Southern University of Science and Technology, Shenzhen 518055, China, and also with the Jiaxing Research Institute, Southern University of Science and Technology, Jiaxing 314031, China. (e-mail: wangjk@sustech.edu.cn, max.meng@ieee.org)

\color{blue}{This is the accepted version of an article published in IEEE Transactions on Systems, Man, and Cybernetics: Systems. Digital Object Identifier 10.1109/TSMC.2026.3736745

© 2026 IEEE. All rights reserved, including rights for text and data mining, and training of artificial intelligence and similar technologies. Personal use is permitted, but republication/redistribution requires IEEE permission. See https://www.ieee.org/publications/rights/index.html for more information.}
}
}
\markboth{IEEE Transactions on Systems, Man, and Cybernetics: Systems}%
{Shell \MakeLowercase{\textit{et al.}}: }


\maketitle

\begin{abstract}
Air-ground collaborative robots have shown great potential in the field of fire and rescue. 
Mapping and navigation, as the key foundation for air-ground collaborative robots to achieve efficient task execution, have attracted a great deal of attention. This growing interest in collaborative robot mapping and navigation is conducive to enhancing the intelligent execution of fire and rescue tasks, but there has been no comprehensive investigation of this field to highlight their strengths. In this paper, we present a systematic review of the air-ground collaborative robots for fire and rescue from a new perspective of mapping and navigation. First, an air-ground collaborative robots framework for fire and rescue missions based on unmanned aerial vehicle (UAV) mapping and unmanned ground vehicle (UGV) navigation is introduced. Then, the research progress of mapping and navigation under this framework is systematically summarized, including UAV mapping, UAV/UGV co-localization, and UGV navigation, with their main achievements and limitations. Based on the needs of fire and rescue missions, the collaborative robots with different numbers of UAVs and UGVs are classified, and their practicality in fire and rescue tasks is elaborated, with a focus on the discussion of their merits and demerits. In addition, the application examples of air-ground collaborative robots in various firefighting and rescue scenarios are given. 
Finally, this paper emphasizes the current challenges and potential research opportunities, providing a reference for practitioners and researchers interested in this rapidly evolving field of air-ground collaborative robotics.
\end{abstract}


\section{Introduction}
In daily life, various emergencies such as fires may occur at any time, threatening the safety of people's lives and property. Fire and rescue, as a vital public safety service, plays an irreplaceable role \cite{bogue2021role, zhang2026distributed}. 
When a fire occurs, firefighters need to rapidly and accurately assess the scene, including fire intensity, spread, and potentially affected areas, and then allocate efforts to firefighting and searching for trapped persons.

With the development of robotics and artificial intelligence, air-ground collaborative robots based on unmanned aerial vehicles (UAVs) and unmanned ground vehicles (UGVs) have been proven to be effective in fire and rescue missions \cite{zhang2023gacf, frering2023multi}. 
They provide more information and operational flexibility, enabling more efficient and intelligent fire and rescue operations \cite{munasinghe2024comprehensive}.
For instance, in fire scene investigation, UAVs can provide high-altitude perspectives and global observations to help firefighters understand the spread of the fire, and the location of the fire source, while UGVs can search on the nearby ground to provide close-range fire monitoring and data collection, especially for complex terrain or areas that require detailed searches. 
In addition, since the operating costs of UAVs and UGVs are relatively low, the air-ground collaborative robots can optimize the utilization of search and rescue resources, as well as reduce the risk of injuries to rescuers \cite{zhang2023weighted, han2024collaborative}. 

For air-ground collaborative robots, mapping and navigation technology, as the key foundation for realizing efficient and autonomous UAV-UGV collaboration, has received increasing attention \cite{miller2024air, liu2022review}. 
In general, with the help of the superior mobility of UAV to quickly perceive large-scale environmental information, on the one hand, a coarse-grained full-scene map for UGV navigation is constructed, and on the other hand, the mission scene is efficiently obtained \cite{wang2024uav}. Based on this, the UGV can achieve rapid path planning and safe navigation to perform fire and rescue missions \cite{munasinghe2024comprehensive, wang2023aerial}. Meanwhile, refined modeling can be achieved in combination with the local information perceived by the UGV to provide more comprehensive environmental information. Therefore, the collaborative paradigm of air-ground robots is a promising solution for improving the efficiency, accuracy, and safety of fire and rescue missions.

Numerous review and survey articles related to air-ground collaborative robots have been presented. Chen \emph{et al}. \cite{chen2015coordination} presented a taxonomy of UAV-UGV cooperative systems and a generalized optimization framework, with the goal of showing how to achieve a common mission goal and perform optimization for specific tasks. Chai \emph{et al}. \cite{chai2024cooperative} discussed the integration of UAV and UGV, focusing on synchronous planning and control mechanisms. Ding \emph{et al}. \cite{ding2021review} reviewed the development of air-ground collaborative systems and identified four key functional roles. Besides, Duan \emph{et al}. \cite{duan2010unmanned} focused on the core issues of heterogeneous collaboration between UAVs and UGVs, and discussed heterogeneous clustering, formation control, formation stability, network control, and practical applications. Liu \emph{et al}. \cite{liu2022review} introduced the basic elements of UAV/UGV collaboration and classified its task types, metrics, and scenarios. 
Guo \emph{et al}. \cite{guo2021service} studied the service collaboration of air-ground collaborative robots to ensure the service quality of service computing. Similarly, Sheng \emph{et al}. \cite{sheng2022space} explored the development and application examples of network technology based on UAVs and UGVs collaboration in the context of high-speed railways. Qin \emph{et al}. \cite{qin2024air} discussed the research progress of air-ground collaborative mobile edge computing. Fei \emph{et al}. \cite{fei2023air} focused on the sensing and communication networks of air-ground collaborative robots and introduced the recent progress of system architecture and protocols. While these surveys provide valuable insights into system architectures, cooperative control, and communication networks, they do not systematically address the critical role of mapping and navigation in enabling effective UAV-UGV collaboration for mission-specific applications such as fire and rescue.

Different from the above existing survey literature, this paper reviews the mapping and navigation methods of air-ground collaborative robots for fire and rescue missions from a new perspective, which is guided by the requirements of UAVs/UGVs to perform fire and rescue operations in a collaborative manner. This perspective highlights the collaborative navigation paradigm enabled by the complementary strengths of UAVs and UGVs in fire and rescue missions, thereby providing a task-oriented understanding of the development of collaborative mapping and navigation technologies in this field. This survey specifically focuses on the paradigm of UAV-enabled mapping for UGV navigation in fire and rescue missions.\footnote{The literature was collected from IEEE Xplore, ScienceDirect, SpringerLink, Web of Science, and Google Scholar, using keywords including “UAV-UGV collaboration”, “air-ground collaborative robots”, “UAV mapping”, “UAV/UGV co-localization”, “UGV navigation”, and “fire and rescue robotics”. Peer-reviewed journal and conference papers closely related to the scope of this survey were prioritized.} Although complementary paradigms, such as UGV-assisted mapping for UAV navigation, have also attracted growing research interest in heterogeneous air-ground systems, they are not comprehensively reviewed in the present survey. 
The contributions of this paper mainly include:
\begin{enumerate}
  \item system architecture of air-ground collaborative robots for fire and rescue, as well as the discussion of the characteristics and applicability of mapping and navigation methods under this framework;
  \item classification and case analyses of air-ground collaborative robots in fire and rescue missions;
  \item discussion on the main research challenges and suggestions for potential research opportunities in the future.
\end{enumerate}

The remainder of this work is organized as follows. Section \ref{section2} introduces the air-ground collaborative navigation framework based on the requirements of performing fire and rescue missions. In Section \ref{section3}, the different types of maps constructed by UAVs for UGV navigation are discussed, which is the basis. Section \ref{section4} summarizes the navigation methods of UGVs from the perspectives of positioning and path planning, while Section \ref{section5} classifies and discusses the collaborative navigation system according to the different numbers of UAVs/UGVs. In Section \ref{section6}, the application of mapping and navigation in fire and rescue scenarios is presented, and specific cases are explained. Finally, the conclusion of the paper and the trends for future research of air-ground collaborative mapping and navigation in the field of fire and rescue are given in Section \ref{section7}.

\section{System architecture of air-ground collaborative robots} \label{section2}
\begin{figure}[!t]
  \centering
  \includegraphics[width=3in]{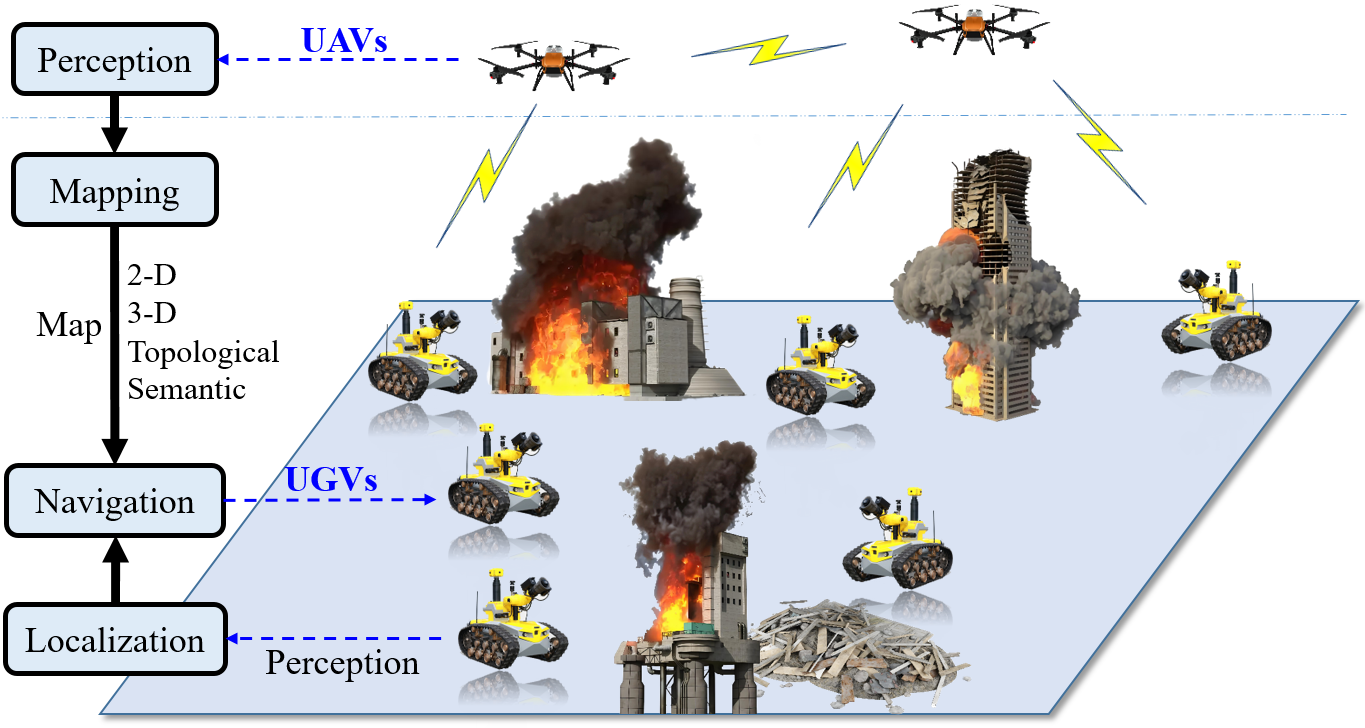}
  \caption{Illustration of air-ground collaborative robots framework for fire and rescue missions based on UAV mapping and UGV navigation.}\label{fig1}
\end{figure}

Air-ground collaborative robots play an increasingly important role in fire and rescue, which can effectively improve the efficiency and safety of fire and rescue operations. As shown in Fig. \ref{fig1}, once a fire occurs, air-ground collaborative robots can quickly enter the fire scenario to perceive the location, scale and environmental conditions of the fire, and perform fire and rescue missions. According to mission requirements, environment mapping and navigation are the key foundations for achieving this goal. Typically, UAVs need to be able to quickly build environmental maps thanks to their large-scale environmental perception capabilities, assisting UGVs in planning and navigation in challenging environments \cite{niu2022unmanned}.

In general, UAVs are equipped with advanced sensors such as high-definition cameras, lidars, and thermal imagers, 
which can quickly obtain real-time situation and heat distribution of disaster areas. With these data, UAVs can build different forms of maps according to the scene characteristics and mission requirements, providing important reference information for UGVs to navigate safely and efficiently \cite{he2024key}. 
Based on this, UGVs plan an optimal path by considering information such as terrain and fire-fighting mission points, and achieve navigation with minimal risk to complete the fire and rescue task \cite{liu2025p}. 
For complex and large-scale rescue scenarios, air-ground collaborative robots can be expanded to multiple UAVs and UGVs to work together to improve the efficiency. For example, multiple UAVs can perceive, map, and monitor different areas, and the collected information is exchanged in real time for coordination and decision-making. Further, according to the task assignment, multiple UGVs use the information perceived by the UAVs to navigate to the assigned task points to collaboratively perform firefighting and rescue tasks.

\begin{figure}[!t]
  \centering
  \includegraphics[width=3in]{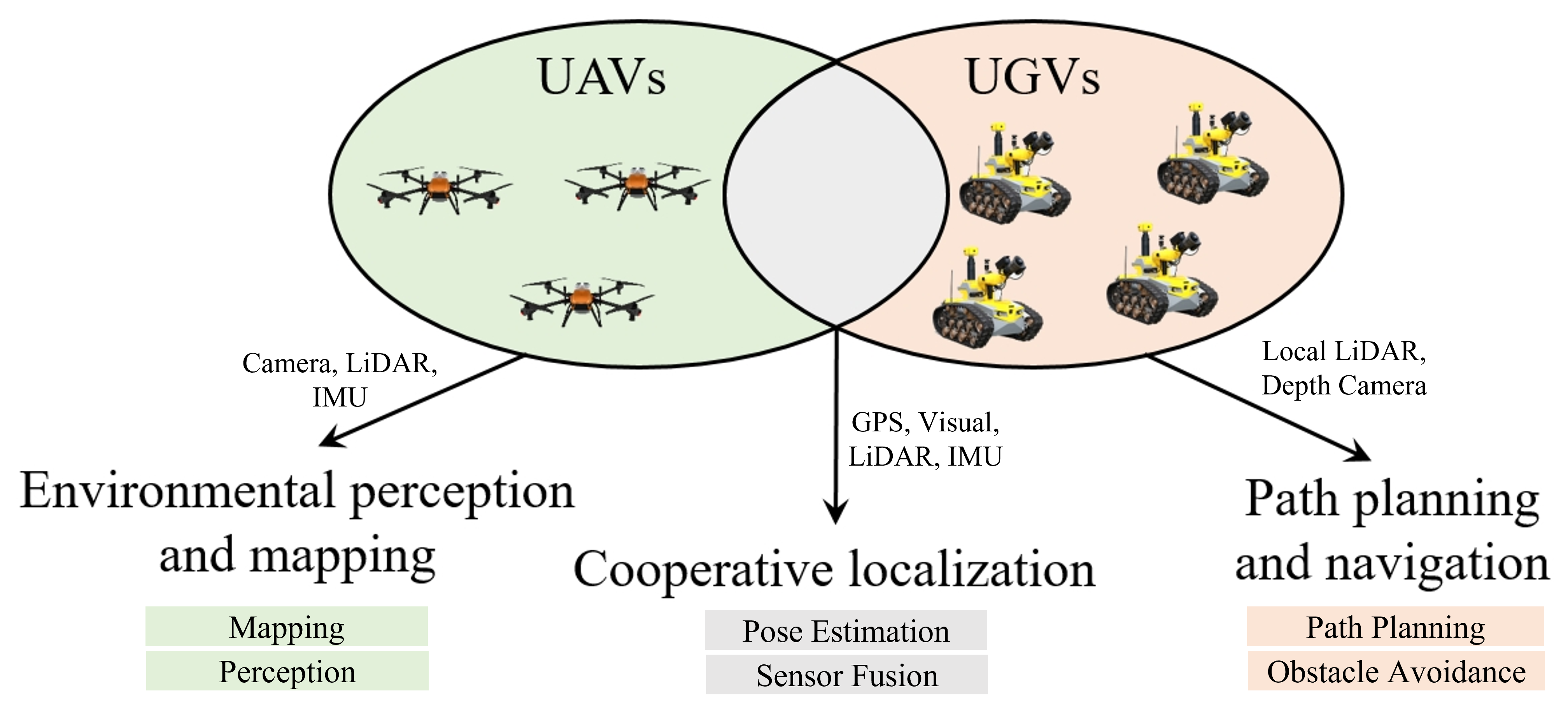}
  \caption{Schematic of the relationship between air-ground collaborative robots for map construction and navigation.}\label{fig2}
\end{figure}

In the above collaborative tasks, this paper mainly focuses on the map construction and navigation of the air-ground collaborative robots, and the schematic of their relationship is shown in Fig. \ref{fig2}. The UAVs are used for rapid environmental perception to build maps, and realize the collaborative localization of the UAVs and UGVs. On this basis, the UGVs perform path planning and navigation with the help of the map constructed by UAVs to complete fire and rescue missions. 

\section{UAV Map Building for UGV Navigation} \label{section3}
Map construction is a prerequisite for robots to complete fire and rescue tasks, providing a basis for robotic motion planning and navigation. In air-ground collaboration, UAVs usually construct maps for UGV navigation due to their superior perception and mobility. To meet UGV navigation requirements, the UAV-constructed maps are typically divided into 2-D grid map, 3-D map, topological map, and semantic map. Each map type exhibits distinct strengths and inherent limitations when applied to fire and rescue scenarios. According to the characteristics of different scenarios, suitable map types can be selected to promote UGV navigation and efficient collaborative operations. Representative works are summarized in Table \ref{table1}.

\begin{table*}[!t]
	\renewcommand{\arraystretch}{1.1}
	\centering
	\scriptsize
	\caption{Summary of representative works on UAV map building for UGV navigation}
	\label{table1}
	\begin{tabular}{
			m{1.4cm}<{\centering}
			m{1.9cm}<{\centering}
			m{3.6cm}<{\centering}
			m{1.5cm}<{\centering}
			m{2.0cm}<{\centering}
			m{2.0cm}<{\centering}
			m{2.8cm}<{\centering}
		}
		\toprule
		\multirow{2.5}*{Map Type} & \multirow{2.5}*{Reference} & \multirow{2.5}*{Feature} & \multicolumn{3}{c}{Experimental Evaluation} & \multirow{2.5}{*}{\parbox{2.8cm}{\centering Limitations in Fire-Rescue scenarios}} \\
		\cmidrule(lr){4-6}
		~ & ~ & ~ & Robots & Sensor & Scenario & ~ \\
		\midrule
		
		\multirow{8}{*}{\centering 2-D Grid Map} 
		& Chatziparaschis \emph{et al}. \cite{chatziparaschis2020aerial} 
		& Constructing a 2-D map, integrated with 3-D OctoMap, to produce a navigable map for UGVs 
		& DJI Matrice 100 
		& Hokuyo LiDAR, Stereo camera 
		& University campus 
		& \multirow{8}{*}{\parbox{2.8cm}{\centering
				2-D grid maps lack explicit vertical and structural information, which limits their effectiveness in multi-level environments or partially collapsed scenes.}} \\
		\cmidrule(lr){2-6}
		& Katikaridis \emph{et al}. \cite{katikaridis2022uav} 
		& Inserting the constructed map into ROS as a Portable Graymap (PGM) image format for efficient UGV navigation 
		& DJI Spreading Wings S1000+ 
		& Sony RX100 III RGB camera 
		& Agricultural environments 
		& \\
		\cmidrule(lr){2-6}
		& Li \emph{et al}. \cite{li2023colag} 
		& Using the UAV occupancy map, the perception-limited UGVs conduct path planning for navigation 
		& Quadrotor 
		& Livox Mid-360 LiDAR 
		& Real-world simulation scenarios
		&  \\
		
		\midrule
		\multirow{8}{*}{\centering 3-D Map} 
		& Delmerico \emph{et al}. \cite{delmerico2017active} 
		& Using UAV for active exploration and employing monocular 3-D reconstruction from low-altitude to map the terrain 
		& Quadrotor MAV 
		& Camera, IMU 
		& Driveway and canyon (Outdoor simulation) 
		& \multirow{7}{*}{\parbox{2.8cm}{\centering
				3-D maps provide richer geometric representations but are sensitive to sensing quality and computational resources, and their reliability may deteriorate under adverse conditions such as smoke, dust, or reduced visibility.}} \\
		\cmidrule(lr){2-6}
		& Arbanas \emph{et al}. \cite{arbanas2018decentralized} 
		& Exploring  unknown environments, constructing 3-D maps by mapping obstacles and walls 
		& AscTec NEO hexacopter 
		& Skybotix VI-Sensor 
		& Narrow corridors, Maze 
		&  \\
		\cmidrule(lr){2-6}
		& Forster \emph{et al}. \cite{forster2013air} 
		& Performing monocular dense 3-D reconstruction, using aerial views to enhance ground maps 
		& NanoQuad MAV
		& Camera, IMU 
		& Indoor and outdoor environment 
		&  \\
		
		\midrule
		\multirow{8}{*}{\centering Topological Map} 
		& Wang \emph{et al}. \cite{wang20212d} 
		& Constructing 3-D ESDF from an aerial perspective, generating a lightweight 2-D topological map 
		& - 
		& Intel realsense D400 
		& Computer room and Shed 
		& \multirow{8}{*}{\parbox{2.8cm}{\centering
				Topological maps facilitate efficient large-scale planning; however, their abstract representations may overlook fine-grained obstacles and traversability details that are essential for safe ground robot navigation in cluttered fire scenes.}} \\
		\cmidrule(lr){2-6}
		& Wang \emph{et al}. \cite{wang2020efficient} 
		& UAV autonomous exploration while constructing environmental topological maps 
		& Developed UAV platform 
		& Intel RealSense RGB-D camera 
		& Indoor environment 
		&  \\
		\cmidrule(lr){2-6}
		& Gomez \emph{et al}. \cite{gomez2020hybrid} 
		& Using the Free-area graph and the Topological map to represent information 
		& - 
		& Hokuyo laser sensor, Asus Xtion camera 
		& Simulation environment / Real-world office 
		&  \\
		
		\midrule
		\multirow{9}{*}{\centering Semantic Map} 
		& Yue \emph{et al}. \cite{yue2021collaborative} 
		& Constructing a semantic map with hierarchical cooperative probabilistic semantic mapping 
		& UAV (with Intel NUC 6i7KYK) 
		& ZED stereo camera 
		& Open Carpark, Indoor-Outdoor mixed environment 
		& \multirow{9}{*}{\parbox{2.8cm}{\centering
			Semantic maps further depend on robust perception and recognition performance, which can be unstable in the presence of occlusions and severe visual degradation.}} \\
		\cmidrule(lr){2-6}
		& Bultmann \emph{et al}. \cite{bultmann2023real} 
		& Integrating multi-modal semantic fusion and label propagation techniques to build a semantic map in real-time 
		& DJI Matrice 210 
		& 3-D LiDAR, RealSense D455 
		& Urban environment, Disaster test site 
		&  \\
		\cmidrule(lr){2-6}
		& Miller \emph{et al}. \cite{miller2022stronger} 
		& Using cell-wise semantic object categories as common map representation for UAV-UGV teams 
		& The Falcon 4 quadrotor 
		& Global shutter RGB camera, ZED-F9P GPS 
		& Outdoor environment 
		&  \\
		
		\bottomrule
	\end{tabular}
\end{table*}

\subsection{2-D Grid Map}
The image data obtained by the UAV can be converted into a 2-D grid map through image processing and object recognition technology \cite{ lu2018survey}. 
The grid map divides the area into grid cells, each of which represents a different type of object or obstacle information. The map can be used for UGV path planning, obstacle avoidance and navigation \cite{fredriksson2024voxel}. 

Li \emph{et al}. \cite{li2016hybrid} constructed a ground map by denoising and correcting the UAV-acquired ground images for UGV path planning and navigation. 
Chatziparaschis \emph{et al}. \cite{chatziparaschis2020aerial} used the UAV to establish an occupancy grid map by extracting the UGV-related area, based on which the UGV performs path planning for safe navigation.
Similarly, Katikaridis \emph{et al}. \cite{katikaridis2022uav} converted the environmental information obtained by the UAV into a grid-based map after preprocessing to support UGV path planning.
In addition, Li \emph{et al}. \cite{li2023colag} proposed an air-ground collaborative architecture to solve the autonomous navigation problem of a group of blind UGVs. In this architecture, the UAV is used for environmental perception and grid map construction to guide a group of UGVs to navigate in an unknown environment. 
For air-ground collaboration, the 2-D grid map constructed by UAV has the characteristics of simplicity, ease of generation and update. This map is suitable for UGV navigation and path planning, with low data volume and computing requirements \cite{zhang2022effective}. 
However, for complex operation scenarios, the 2-D map representation capability is limited and cannot provide detailed perceptual information such as height and semantics, which makes it difficult for UGVs to adapt to complex terrain. 


\subsection{3-D Map}
3-D maps provide UGVs with a stereoscopic view of the ground surface, including information such as terrain, buildings, trees, and other height features \cite{xie2023circular}. 
Typically, 3-D maps are represented in the form of point cloud maps, octree maps, etc. These maps can be used for high-precision positioning, obstacle avoidance, and path planning for UGVs, and are particularly suitable for complex environments or rugged terrain.

For example, Delmerico \emph{et al}. \cite{delmerico2017active} proposed an air-ground collaborative active exploration method that uses a UAV to perceive and generate a 3-D map of the environment to provide the UGV with traversable paths through unknown terrain. 
Arbanas \emph{et al}. \cite{arbanas2018decentralized} built a 3-D OctoMap by UAV exploration and projected it onto the ground to guide the UGVs.
Similarly, Wang \emph{et al}. \cite{wang2021intelligent} developed a ground-air collaborative navigation system that combines an OctoMap-based 3-D environment mapping method for map creation in complex environments. Moreover, Fedorenko \emph{et al}. \cite{fedorenko2018global} constructed a 3-D point cloud of the environment using a UAV and then investigated the feasibility of using this point cloud map directly for UGV planning and navigation. When the initial relative attitude of UAV and UGV is unknown, Pei \emph{et al}. \cite{pei2023air} used only a laser scanner to achieve air-ground collaborative point cloud fusion and 3-D map construction. For air-ground collaboration, Forster \emph{et al}. \cite{forster2013air} used a monocular camera carried by a UAV to construct a dense 3-D point cloud map to enhance the efficiency of UGV positioning and navigation. 
Hence, 3-D maps provide rich spatial information and a more realistic and accurate representation of the environment, supporting the effective operation of air-ground collaborative robots in complex scenarios \cite{zhang2021safe, tang2023perception}.
However, they require large storage and computation, and more complicated to acquire and update than 2-D maps.


\subsection{Topological Map}
As an abstract map representation, topological maps are usually composed of nodes and edges, where nodes represent key locations in the environment and edges represent direct or indirect connections between nodes \cite{zhang2022building, wang2023ternformer}. 
In air-ground collaboration, UAVs can generate road-network maps based on topological relationships, simplifying the road network representation for UGV path planning and navigation.

Wang \emph{et al}. \cite{wang20212d} proposed a method for constructing a two-dimensional topological map of the ground environment from an aerial perspective using UAVs. This method can be effectively applied to global path planning for UGVs. 
Wang \emph{et al}. \cite{wang2020efficient} used the depth camera carried by the UAV to build a topological map that is completed incrementally as the UAV flies. 
This map explicitly shows the topological structure of the 3-D environment, and can also effectively provide information gain and cost-to-go for the candidate areas to be explored.
To improve the efficiency of global planning, Blochliger \emph{et al}. \cite{blochliger2018topomap} put forward Topomap, a framework for simplifying navigation tasks, by converting sparse feature maps into 3-D topological maps. Gomez \emph{et al}. \cite{gomez2020hybrid} developed a hybrid mapping method that combines the efficiency of topological maps with the accuracy of dense representations for path planning. Besides, Zhang \emph{et al}. \cite{zhang2023gacf} proposed a cross-domain collaborative unmanned framework called GACF. This framework utilizes a topological construction algorithm based on aerial perspective images from UAVs for road network extraction, providing a global traversable map for UGV path planning purposes. 
Such topological maps are suitable for areas with relatively simple road networks, such as highways and urban grids \cite{feng2021topology}, but their lack of detailed geometric information limits their use in complex environments such as fire and rescue scenarios.


\subsection{Semantic Map}
Semantic maps provide richer environmental perception and scene understanding for air-ground collaborative robots \cite{zhao2024enhanced}. 
In this context, UAVs can use visual sensors to build semantic maps containing geographical elements such as roads, buildings, and traffic signs to support high-level decision-making and behavior planning of UGVs.

Vasi{\'c} \emph{et al}. \cite{vasic2020deep} performed semantic segmentation on the images captured from the UAV's viewpoint using deep neural networks, thereby providing effective path planning for successful navigation of the UGV. Maturana \emph{et al}. \cite{maturana2018real} proposed a semantic mapping system for automatic off-road driving of UGVs, which provides a richer environmental representation than pure geometric maps. 
Furthermore, Yue \emph{et al}. \cite{yue2021collaborative} established a hierarchical collaborative probabilistic semantic mapping framework by combining information from UAVs and UGVs to obtain semantic point clouds, which are then utilized for generating semantic maps. Bultmann \emph{et al}. \cite{bultmann2023real} designed a multi-sensor model for real-time semantic reasoning and fusion of UAV systems. 
It is capable of fusing and generating a global semantic map, holding tremendous potential in fast autonomous or remote-controlled semantic scene analysis. This system can be applied to disaster inspections. Miller \emph{et al}. \cite{miller2022stronger} exploited an integrated air-ground collaboration system where a semantic map created in real time by the UAV enables UGVs to perform localization, planning, and navigation. The system is demonstrated to be effective in simulating disaster scenarios. 
By emphasizing the semantic and attribute information of geographic features, semantic maps are particularly suitable for complex urban environments \cite{shang2026uav}.
However, semantic maps still face notable challenges in fire and rescue scenarios, particularly in handling ambiguous and dynamic features caused by smoke, flames, occlusions, and rapidly changing environments. These factors may lead to unstable semantic perception and misclassification of objects, which can degrade the reliability of task planning and navigation. 


Unlike generic environments where map selection is primarily driven by representation accuracy and computational efficiency, fire and rescue scenarios introduce unique constraints, including degraded perception caused by smoke and occlusions, dynamic fire propagation requiring rapid updates, and partially collapsed structures requiring 3-D geometric and structural information. These factors shift mapping priorities toward robustness to sensing degradation, real-time adaptability, and task-oriented perception. Different map representations exhibit distinct characteristics and applicability. For example, 2-D grid maps are advantageous for time-critical deployment due to their low computational burden, whereas 3-D maps are more suitable for collapsed structures, uneven terrain, or multi-level environments requiring detailed geometric information. Topological maps facilitate efficient global guidance in large-scale or GPS-denied environments, while semantic maps provide richer scene understanding to support high-level decision-making, such as victim search and hazard identification. Accordingly, integrating multiple map representations can provide more comprehensive and reliable support for navigation in fire and rescue missions.

\section{UAV Map-Based UGV Navigation} \label{section4}
In the navigation of air-ground collaborative robots, the core links include environment mapping, localization and path planning. For fire and rescue tasks, after the UAV creates the corresponding map according to the task, the UGV uses its own sensors to identify the surrounding environmental features based on these maps to achieve positioning. Subsequently, the UGV plans a safe and efficient route to ensure that the fire and rescue mission can be successfully completed. Map construction has been explained in Section \ref{section3}, and localization and path planning will be discussed below.

\subsection{UAV and UGV Co-Localization}
Co-Localization is an indispensable component of air-ground collaborative robot navigation, and its accuracy directly affects the success rate and efficiency of task execution. Typically, localization methods can be divided into GPS/GNSS localization, inertial navigation system (INS) localization, visual localization, lidar localization, and multi-sensor fusion localization, as shown in Table \ref{table2}.

\begin{table*}[!t]
	\renewcommand{\arraystretch}{1.1}
	\centering
	\caption{Summary of Representative Works on UAV and UGV Co-Localization}\label{table2}
	\scriptsize
	\renewcommand\arraystretch{1.1}
	\begin{tabular}{
			m{1.5cm}<{\centering} 
			m{2.1cm}<{\centering} 
			m{4.5cm}<{\centering} 
			m{4.5cm}<{\centering} 
			m{3.5cm}<{\centering}
		}
		\toprule
		\multirow{1}*{Localization Methods} & \multirow{1}*{Reference} & \multirow{1}*{Merit} & \multirow{1}*{Demerit} & \multirow{1}*{Fire-Rescue Limitations} \\
		\midrule
		\multirow{4}*{\parbox{2.1cm}{\centering GPS/GNSS \\localization}}
		& Li \emph{et al}. \cite{li2023consistent} 
		& Positioning through information sharing and consistency checks 
		& Not delved deeply into the integrity algorithm issues for inter-vehicle and multi-sensor integration in multi-robot systems 
		& \multirow{4}{*}{\parbox{3.5cm}{\centering May fail in indoor, underground, or structurally damaged environments due to signal blockage; Unreliable in dense urban canyons with limited satellite visibility}} \\
		\cmidrule(lr){2-4} 
		~ & Correa-Caicedo \emph{et al}. \cite{correa2021gps} 
		& Having some competitiveness in GPS data correction; With relatively low dependence on parameters and sensors 
		& Limited generalization capability, need retraining in new geographic regions to fit the data
		& \\
		\cmidrule(lr){1-5} 
		\multirow{3}*{\parbox{2.1cm}{\centering INS \\localization}} 
		& Yilmaz \emph{et al}. \cite{yilmaz2023novel} 
		& Exploring for GPS-denied environments 
		& Lack of validation of robustness in real scenarios
		& \multirow{3}{*}{\parbox{3.5cm}{\centering Accuracy degradation over prolonged rescue operations due to accumulated drift; Sensitivity to vibration and thermal variations in fire environments}} \\
		\cmidrule(lr){2-4} 
		~ & Zhou \emph{et al}. \cite{zhou2023wi} 
		& Providing attitude information and offering continuous measurement data (INS)
		& Accumulated IMU errors reducing localization accuracy
		& \\
		\cmidrule(lr){1-5}
		\multirow{5}*{\parbox{2.1cm}{\centering Visual \\localization}} 
		& Zhang \emph{et al}. \cite{zhang2022factor} 
		& Taking into account the unstructured and feature-deficient nature of the environment 
		& Insufficient features may lead to inaccurate positioning or failure
		& \multirow{3}{*}{\parbox{3.5cm}{\centering Vulnerability to smoke, flames, low visibility, and dynamic scene changes; Performance degradation under severe occlusion and visual degradation in fire environments}} \\
		\cmidrule(lr){2-4} 
		~ & Song \emph{et al}. \cite{song2024Robotl} 
		& Combining semantic information for localization; Considering the relocalization efficiency in dynamic indoor environments 
		& Coarse positioning accuracy depends on detection precision
		& \\
		\cmidrule(lr){1-5}
		\multirow{4}*{\parbox{2.1cm}{\centering Lidar \\localization}} 
		& Lima \emph{et al}. \cite{de2023air} 
		& Fusing aerial and ground perception data; Exploring the positioning issue in outdoor unstructured environments without GPS 
		& Lack of evaluation of the impact of environmental appearance changes on the robustness of positioning
		& \multirow{4}{*}{\parbox{3.5cm}{\centering Possible performance degradation due to airborne particulates (e.g., smoke and dust); Constrained by high computational and memory requirements}} \\
		\cmidrule(lr){2-4} 
		~ & Caballero and Merino \cite{caballero2021dll} 
		& Based on raw point cloud data; Evaluating the robustness under odometry errors 
		&  Low spatial efficiency; Large memory required
		& \\
		\cmidrule(lr){1-5}
		\multirow{4}*{\parbox{2.1cm}{\centering Multi-sensor\\fusion\\localization}} 
		& Kayhani \emph{et al}. \cite{kayhani2022tag} 
		& Based on camera and IMU; Considering the cost-effectiveness and lightweight design of the localization method 
		& Dependence on manual tag deployment and maintenance
		& \multirow{4}{*}{\parbox{3.5cm}{\centering Robustness improvement via multi-sensor fusion; Sensitivity to inter-sensor inconsistency under heterogeneous disturbances}} \\
		\cmidrule(lr){2-4}
		~ & Patoliya \emph{et al}. \cite{patoliya2022robust} 
		& Integrating GPS and LiDAR data; Considering error accumulation in an unbounded environment 
		&  Limited in mapping large open space; Insufficient consideration for use on non-planar surfaces
		& \\
		\bottomrule
	\end{tabular}
\end{table*}

\subsubsection{GPS/GNSS localization}
GPS or GNSS can provide basic coordinate information for UAVs and UGVs for rough localization \cite{chen2023milestones}. 
Based on this, collaborative robots use the localization information in the same coordinate system to perform coordinate transformation to allow them to use a common map for navigation. Li \emph{et al}. \cite{li2023consistent} designed a GNSS-based collaborative localization method that uses differential pseudo-range and carrier phase observations to obtain the accurate relative position between robots. 
In addition, the GPS-based localization solution is very sensitive to the environment in which the robot is located, and it may experience system failures due to environmental conditions. To solve this problem, Correa-Caicedo \emph{et al}. \cite{correa2021gps} proposed an intelligent system based on fuzzy logic that obtains information from sensors and corrects the absolute position of the robot based on longitude and latitude. 
While GPS-based localization offers wide coverage and is suitable for outdoor open environments, it fails in GPS-denied scenarios such as indoor areas or dense urban fire scenes where signals are easily obstructed \cite{shimizu2023accuracy}.


\subsubsection{INS localization}
INS measures the acceleration of the carrier in the inertial reference frame and integrates it over time. The resulting information is transformed into the navigation coordinate system to obtain the carrier's velocity, yaw angle, and position for robot localization \cite{engelsman2023information}.
Yilmaz \emph{et al}. \cite{yilmaz2023novel} designed an INS-based localization method in which the INS provides reference coordinates, and the positions of individual modules are computed using the Denavit-Hartenberg method. 
Traditional INS-based localization has poor accuracy and severe drift due to error accumulation. To solve this problem, Zhou \emph{et al}. \cite{zhou2023wi} proposed a tightly integrated localization system based on Wi-Fi round-trip time, encoder, and inertial measurement unit (IMU). Through the error state Kalman filter, the inertial information from the IMU is fused with the measurement results from the encoder to suppress the accumulated error of INS. 
Notably, INS does not rely on external signals. Moreover, INS can work in a variety of environments such as air, ground, and underwater, and is not restricted by weather and lighting conditions \cite{cohen2024inertial}. 
However, INS is sensitive to vibration and temperature changes, which may degrade localization accuracy, and its alignment time can be long in complex environments.

\subsubsection{Visual localization}
For air-ground collaborative robots, vision-based localization typically involves UAVs using visual sensors to collect environmental data and construct a scene map, after which UGVs match surrounding environmental information with the scene map and database to achieve localization \cite{zhang2026long}. 
Zhang \emph{et al}. \cite{zhang2022factor} suggested a visual localization method that adopts fiducial markers and factor graphs to achieve reliable localization performance. In this method, fiducial markers are introduced into the front-end module, and then the detection algorithm AprilTag2 is used to identify the markers and extract feature points to locate the robot. The visual localization method can solve the problem of localization and navigation of air-ground collaborative systems without GPS \cite{cheng2023unmanned}. It can model scene maps based only on visual sensors, and realize robot localization through feature matching and map registration \cite{song2024Robotl}. 
However, its performance may deteriorate under insufficient lighting or significant environmental changes, and it typically requires substantial computational resources.

\subsubsection{Lidar localization}
Lidar localization is to align the geometric feature data collected by the lidar sensors with the map of the surrounding environment for air-ground collaborative robot localization \cite{de2023air}. 
Caballero and Merino \cite{caballero2021dll} introduced a direct lidar localization scheme. This method uses a point cloud to map registration based on nonlinear optimization of the distance between the points and the 3-D map for robot localization. 
Such a solution can achieve robot localization and navigation in areas without GPS, but the memory required to build the 3-D map is very large. To this end, Zhang \emph{et al}. \cite{zhang2022design} designed a dual-lidar high-precision natural navigation system based on the ROS platform. The system adopts the Lidar-SLAM method based on graph optimization to construct a 2-D environmental map, and then uses the PF algorithm in MRPT for robot localization. Lidar-based technology is very effective in detecting obstacles and accurately measuring distances \cite{chalvatzaras2023survey}. It is not affected by lighting conditions and is suitable for use in a variety of scenarios. 
However, the cost of lidar is relatively high, and in some cases it may be affected by bad weather such as dust and rain.
%

\subsubsection{Multi-sensor fusion localization}
The multi-sensor fusion localization method can effectively address the limitations of a single sensor, enabling the robot to achieve more accurate localization and navigation in different environments and conditions \cite{yin2024tightly}. 
Donati \emph{et al}. \cite{donati20223d} designed a method based on GPS, IMU and ultrasound sensors. The method combines the data provided by multiple sensors with the information provided by 3-D low-complexity maps to support the localization of air-ground collaborative robots. Kayhani \emph{et al}. \cite{kayhani2022tag} proposed a tag-based visual inertial localization method, which takes advantage of the vision to identify the tag position and uses the manifold EKF to fuse inertial data and tag measurements for UAV localization. In addition, Patoliya \emph{et al}. \cite{patoliya2022robust} developed a robot localization system that integrates GPS and LiDAR data. Similarly, Yousuf \emph{et al}.  \cite{yousuf2016sensor} introduced a method for robot localization that fuses INS, GPS and odometer sensors. Furthermore, recent studies have explored millimeter-wave radar and vision fusion for robust localization in challenging environments. Wang \emph{et al}. \cite{wang2024millimeter} fused radar and visual data in a semantic SLAM framework to enhance point cloud registration and loop closure reliability. Zhu and Yang \cite{zhu2025rdynaslam} incorporated 4-D radar into visual SLAM to mitigate the influence of dynamic objects during pose estimation. 
Multi-sensor fusion therefore provides complementary sensing information for maintaining localization accuracy and robustness in challenging and dynamically changing environments \cite{li2026localization}.

From the perspective of fire and rescue applications, different co-localization techniques face distinct challenges posed by mission-specific environmental factors. GNSS-based methods may fail in indoor, underground, or structurally damaged environments due to signal blockage. Visual localization is vulnerable to smoke, low visibility, and dynamic scene changes, while LiDAR performance can be degraded by airborne particulates and thermal disturbances. Although INS provides self-contained localization capability, its accuracy may deteriorate over prolonged operations due to accumulated drift. Consequently, enhancing robustness under sensing degradation has become a key requirement, making multi-sensor fusion an increasingly important strategy for reliable localization in fire and rescue missions.

\subsection{UGV Navigation}
Efficient navigation of UGV is a key step for air-ground collaborative robots to complete firefighting and rescue missions. To achieve this, UGV needs to be able to select the appropriate navigation method based on the map constructed by the UAV. In such context, according to the type of map established by the UAV, the UGV navigation method is classified into navigation based on 2-D grid map, 3-D map, topological map, and semantic map. A summary of the representative works is also presented in Table \ref{table3}.

\begin{table*}[!t]
	\renewcommand{\arraystretch}{1.3}
	\centering
	\caption{Summary of Representative Works on UGV Navigation Based on UAV Maps}\label{table3}
	\scriptsize
	\renewcommand\arraystretch{1.1}
	\begin{tabular}{
			m{1.5cm}<{\centering} 
			m{1.4cm}<{\centering} 
			m{4.5cm}<{\centering} 
			m{1.2cm}<{\centering} 
			m{2.3cm}<{\centering} 
			m{2.2cm}<{\centering}
			m{2.3cm}<{\centering}
		}
		\toprule
		\multirow{2.5}*{\parbox{1.5cm}{\centering Applied Map\\Types}} & \multirow{2.5}*{Reference} & \multirow{2.5}*{Feature} & \multicolumn{3}{c}{Experimental Evaluation} & \multirow{2.5}{*}{Fire-Rescue Limitations} \\
		\cmidrule(lr){4-6} 
		~ & ~ & ~ & Sensor & Path Planning Method & Scenario & ~ \\
		\midrule
		\multirow{5}*{\parbox{1.2cm}{\centering 2-D grid\\map-based}}
		& Zhang \emph{et al}. \cite{zhang2021user} 
		& Based on the occupancy grid map, combined with virtual areas, performing global and local path planning for human-aware indoor navigation 
		& SICK TIM561 laser rangefinder 
		& D* algorithm / DWA 
		& Indoor environment 
		& \multirow{5}{*}{\parbox{2.3cm}{\centering Limited adaptability to 3-D obstacles and dynamic fire scenes; Dependence on map accuracy}} \\
		\cmidrule(lr){2-6} 
		~ & Zhang \emph{et al}. \cite{zhang2024bi} 
		& Constructing a 2-D grid map and using Bidirectional RRT with assisting metric for path planning to achieve indoor navigation 
		& Laser Radar 
		& Bi-AM-RRT* 
		& Indoor environment 
		& \\
		\cmidrule(lr){1-7} 
		\multirow{5}*{3-D map-based} 
		& Delmerico \emph{et al}. \cite{delmerico2017active} 
		& Using UAV monocular 3-D reconstruction to create topographic maps, UGVs calculate paths for navigation 
		& Rangefinder, Camera, IMU 
		& D* algorithm (Modified version) 
		& Driveway and Canyon (Outdoor simulation)
		& \multirow{5}{*}{\parbox{2.3cm}{\centering High computational cost; Reduced navigation robustness under sensing degradation}} \\
		\cmidrule(lr){2-6} 
		~ & Arbanas \emph{et al}. \cite{arbanas2018decentralized} 
		& Utilizing the map constructed by UAV for path planning, UGV accomplishes navigation through the cooperation of UAV 
		& VI-Sensor 
		& Lattice-based algorithm 
		& Narrow corridors, Maze
		& \\
		\cmidrule(lr){1-7}
		\multirow{4}*{\parbox{1.5cm}{\centering Topological\\map-based}} 
		& Wang \emph{et al}. \cite{wang20212d} 
		& Using a 2-D topological map constructed by UAVs, UGVs perform global path planning for navigation 
		& Lidar, Camera 
		& A* search 
		& Computer room and Shed / Simulation environment
		& \multirow{4}{*}{\parbox{2.3cm}{\centering Difficulty in representing obstacles and feasible paths in cluttered fire environments; Challenges for safe navigation}} \\
		\cmidrule(lr){2-6}
		~ & Zuo \emph{et al}. \cite{zuo2023real} 
		& Constructing reduced approximated generalized Voronoi graph, employing a three-phase rapid path search method for 3-D space navigation 
		& Velodyne VLP-16 laser scanner 
		& Three-stage grid-graph-combined approach / Dijkstra 
		& Underground parking, Warehouse and Treelawn, Factory
		& \\
		\cmidrule(lr){1-7}
		\multirow{6}*{\parbox{1.5cm}{\centering Semantic\\map-based}} 
		& Kostavelis \emph{et al}. \cite{kostavelis2016robot} 
		& Integrating semantic mapping with robotic navigation strategies to construct semantic maps and achieve hierarchical navigation 
		& Camera 
		& Dijkstra algorithm 
		& COLD dataset (Ljubljana section)
		& \multirow{6}{*}{\parbox{2.3cm}{\centering Reduced semantic perception accuracy in complex fire scenes; Degraded navigation performance due to reduced semantic map quality}} \\
		\cmidrule(lr){2-6}
		~ & Deng \emph{et al}. \cite{deng2020semantic} 
		& Integrating semantic segmentation and SLAM frontend to generate dense semantic maps, using semantic information to enhance path planning algorithms 
		& Realsense D435 camera 
		& Algorithm incorporating semantic information 
		& RRL competition environment
		& \\
		\bottomrule
	\end{tabular}
\end{table*}

\subsubsection{2-D grid map-based navigation}
2-D grid maps can be directly used for UGV navigation, allowing the UGV to plan the optimal path from the current position to the target position \cite{zhang2021user}. Dijkstra \cite{dijkstra1959note} proposed an algorithm for solving the single-source shortest path problem in graphs with nonnegative edge weights. 
To improve the computational efficiency of Dijkstra's algorithm, Hart \emph{et al}. \cite{hart1968formal} developed the A* algorithm, which uses a heuristic function to prioritize nodes closer to the target. This enables the robot to find paths more efficiently. 
In response to the problems of long calculation time and many search nodes in A*-based global path planning, a variety of improved algorithms based on search and sampling are proposed for UGV planning and navigation \cite{zhang2024bi, ma2026unmanned}. 
Moreover, recent learning-based planners have been explored to improve global path planning efficiency and adaptability in complex and dynamic environments, showing promising potential compared with classical RRT and A* methods \cite{hu2026review, qureshi2020motion}. However, their application to fire and rescue missions remains challenging due to the reliance on task-specific training data, limited generalization from simulation to real disaster environments, and substantial computational requirements \cite{aradi2020survey}. Although the UGV navigation method based on 2-D grid can achieve efficient and safe navigation, its navigation performance is subject to the accuracy of the map, and it has limitations such as poor adaptability to 3-D obstacles and highly dynamic scenes where fires spread rapidly.


\subsubsection{3-D map-based navigation}
3-D maps are generally not used directly for UGV navigation. They are usually converted into 2-D or 2.5-D maps for navigation. 
Of these, 2.5-D maps are particularly suitable for navigation \cite{zhang2024Environment}, as they enable efficient 2-D path planning while retaining terrain height information. 
For example, Delmerico \emph{et al}. \cite{delmerico2017active} combined the altitude and terrain data of the 3-D map and used the D* algorithm to plan a navigable path for the UGV that takes into account both the difficulty of the terrain and the shortest completion time, thereby improving the navigation efficiency in complex environments. 
Analogously, Basso \emph{et al}. \cite{basso2024sharing} designed a 3-D map sharing strategy for multi-UAV systems, in which each UAV contributes its locally constructed 3-D map to the cooperative team, and the globally merged map can also be reused for UGV navigation. Arbanas \emph{et al}. \cite{arbanas2018decentralized} projected the 3-D OctoMap onto the ground for UGV navigation with a lattice-based path planning algorithm. Weerakoon \emph{et al.} \cite{weerakoon2023graspe} leveraged a graph neural network to fuse RGB, 3-D LiDAR, and odometry information for trajectory traversability prediction, enabling robust navigation under partial sensor degradation. The advantage of navigation based on 3-D maps is that it can provide richer environmental information for UGVs, while its high computational cost and sensitivity to sensing quality limit its robustness in smoke-filled and rapidly changing fire scenes.


\subsubsection{Topological map-based navigation}
UGV can directly use the topological map constructed by UAV to plan the reference line for navigation, which can effectively reduce the path planning time \cite{zhang2022building}. 
Blochliger \emph{et al}. \cite{blochliger2018topomap} proposed Topomap, a framework for creating multifunctional topological maps and reliable navigation, which uses graph search to traverse topological nodes to obtain the shortest path. 
Wang \emph{et al}. \cite{wang20212d} generated 2-D topological maps using UAV observations, on which global paths for UGV navigation can be efficiently planned. 
Oleynikova \emph{et al}. \cite{oleynikova2018sparse} constructed a sparse graph containing 3-D topological information for robot planning and navigation in a 3-D environment. 
Ravankar \emph{et al}. \cite{ravankar2017hybrid} proposed a hybrid topological mapping and navigation method for mobile robots, using the A* algorithm to perform navigation between topological nodes. Zuo \emph{et al}. \cite{zuo2023real} introduced the Voronoi diagram into the topological mapping model. On this basis, a three-stage fast path search and navigation method was designed. Topological maps allow UGVs to perform efficient path planning and navigation. However, due to the large amount of debris in fire and rescue scenarios, it is difficult for topological maps to accurately represent obstacles and feasible paths, making it challenging for UGVs to safely navigate based on topological maps.


\subsubsection{Semantic map-based navigation}
Semantic maps provide UGVs with richer environmental information, such as scene and object semantics, as well as traversable and non-traversable areas, enabling UGVs to better perceive and understand their surrounding environment \cite{xia2020survey}. 
Kostavelis \emph{et al}. \cite{kostavelis2016robot} constructed a multi-layer semantic map incorporating spatial and temporal information to provide robots with a hierarchical navigation solution. 
Deng \emph{et al}. \cite{deng2020semantic} designed a semantic SLAM framework for rescue robot navigation, enabling robots to identify different terrain types in complex environments and select routes with better traversability. Vasi{\'c} \emph{et al}. \cite{vasic2020deep} used a deep neural network to segment traversable areas in UAV images and proposed a semantic-map-based UGV path planning method for dynamic environments. 
Semantic maps support semantic planning and navigation for UGVs. However, their creation and updating require substantial computational resources \cite{badrloo2022image}. 
In complex fire and rescue environments, accurately identifying and segmenting objects, areas, and other semantic elements can be challenging, affecting mapping quality and consequently UGV navigation performance.


In fire and rescue missions, co-localization and navigation face challenges including sensor degradation due to smoke and thermal disturbances, as well as rapidly changing scenes caused by fire spread and structural collapse. Compared with generic environments, these scenarios place greater emphasis on robustness to sensing degradation and responsiveness to dynamic changes. Multi-sensor fusion is commonly adopted to improve localization reliability, while lightweight and reactive planning strategies facilitate rapid adaptation.

\section{Classification of Air-Ground Collaborative Robots in Fire and Rescue Missions} \label{section5}
For fire and rescue missions, the number of UAVs and UGVs in the air-ground collaborative robot system can vary depending on task requirements, environmental conditions, and available resources. In this context, the composition of air-ground collaborative systems can generally be classified into four categories: a single UAV with a single UGV, a single UAV with multi-UGVs, multi-UAVs with a single UGV, and multi-UAVs with multi-UGVs. 
These configurations are not merely distinguished by agent count. Instead, the proposed categorization is grounded in the distinct modes of collaboration for mapping and navigation inherently associated with each team composition. 
From the perspectives of mapping and navigation, some representative studies have been validated in generic applications, while their underlying collaboration mechanisms provide valuable insights for the application of air-ground collaborative systems in fire and rescue missions.
Specifically, these mechanisms include UAV-assisted mapping and perception, leader-follower coordination, one-to-many information sharing, collaborative mapping and perception with multiple UAVs, and distributed coordination among multiple UAVs and UGVs. However, generic applications often assume relatively stable environments and reliable sensing and communication, which may not hold in fire and rescue scenarios due to rapid environmental changes, sensing degradation, and communication constraints. Practical deployment of these mechanisms therefore requires adaptation to the specific sensing, mapping, navigation, and coordination requirements of fire and rescue missions.

\subsection{A single UAV with a single UGV}
For small-scale fire monitoring and rescue missions, a single UAV paired with a single UGV is a widely adopted configuration. In such a system, the UAV provides aerial surveillance and builds a map of the environment, while the UGV navigates using this map to perform specific tasks. Compared with configurations involving multiple UAVs or UGVs, the roles of the two platforms are clearly divided, resulting in relatively simple localization and navigation.

Giakoumidis \emph{et al}. \cite{giakoumidis2012pilot} designed a heterogeneous architecture consisting of a UGV and a UAV, where the UAV serves as a separable remote vision system for the UGV. In the system, the environmental images captured by the UAV are stitched to construct a 2-D map, and then the FM2 algorithm is used to implement UGV path planning. This framework demonstrates the potential of air-ground collaboration for fire surveillance and rescue in unknown environments. 
Similarly, Fankhauser \emph{et al}. \cite{fankhauser2016collaborative} developed a collaborative navigation framework for flying and walking robots. The method uses the UAV’s onboard monocular camera to create a visual feature map for localization and a dense elevation map of the terrain. Based on this shared model, the UGV localizes itself and plans its path by interpreting the elevation map in terms of traversability. 
In order to enable air-ground collaborative robots to perform autonomous collaborative tasks in disaster scenarios, Mueggler \emph{et al}. \cite{mueggler2014aerial} took advantage of the UAV to capture ground images for 2-D grid map construction, and employed the Dijkstra algorithm for UGV path planning. However, each action of the UGV is iteratively directed by the UAV, resulting in low autonomy. In addition, Christie \emph{et al}. \cite{christie2017radiation} proposed a UAV-UGV cooperative solution for task execution using scene understanding. 
In this solution, a UAV is used to identify points of interest, model digital elevation of the unknown scene, and build a 3-D semantic map based on semantic segmentation. Then based on these outputs, the UGV utilizes its onboard LiDAR for collision-free path planning and autonomous task execution. 
While this single-UAV-single-UGV configuration has been successfully applied, it still faces limitations in task execution capability and adaptability to complex, large-scale fire and rescue scenarios \cite{mondal2024robust}.


\subsection{A single UAV with multi-UGVs}
The collaboration of a UAV with multi-UGVs can improve the efficiency of mission execution and provide faster response capabilities. In this configuration, the UAV can provide comprehensive environmental information to UGVs, allowing UGVs to collaboratively perform search and rescue missions \cite{jin2024deep}. In addition, UGVs can also share map data, sensor information, etc. to improve collaborative work efficiency.

Miller \emph{et al}. \cite{miller2022stronger} proposed a semantic-based air-ground collaborative robot system framework that can realize collaborative navigation of a UAV and multiple UGVs. The system deploys a UAV to create a semantic map of the ground in real time, which is then used for UGVs' positioning and semantic navigation tasks. 
For efficient reconnaissance by multi-UGVs in unknown environments, Zhang \emph{et al}. \cite{zhang2023formation} proposed a formation cooperative reconnaissance strategy. In this strategy, a frontier-based exploration method is used to identify reconnaissance targets in unknown areas, and the A* algorithm is then employed to plan the exploration order and paths through these targets. 
Similarly, Zang \emph{et al}.  \cite{zang2024coordinated} designed a multi-UGVs system solution based on coordinated behavior planning and trajectory planning. However, the coordination problem between the UAV and multiple UGVs is not considered. In addition, in fire and rescue missions, UGVs may be scattered over a large area, which may affect their collaboration due to the limited perception capabilities. In this case, the UAV not only needs to perform map construction to assist UGV navigation, but also needs to act as a messenger to support indirect communication of UGV. Ding \emph{et al}. \cite{ding2017path} addressed the problem of path planning for a messenger UAV in a system involving aerial and ground collaboration with multiple UGVs. This provides a solution for achieving one-to-many air-ground collaborative navigation, facilitating real-time information sharing and coordinated rescue operations in fire and rescue scenarios. To solve the path planning problem of collaborative heterogeneous robots consisting of a UAV and multiple UGVs, Ding \emph{et al}. \cite{ding2022memetic} developed an effective memetic algorithm to find the shortest route that allows the messenger UAV to access all moving UGVs. This treatment allows the UGVs to perform fire and rescue missions in a collective manner over a large area, and improve mission efficiency to a certain extent. However, in large-scale and complex fire and rescue scenarios, a single-UAV-multi-UGV configuration requires sustained perception and modeling capabilities, making the UAV subject to endurance limitations, increased communication load, and coordination bottlenecks as the team size grows.


\subsection{Multi-UAVs with a single UGV}
The configuration of multi-UAVs with a single UGV is suitable for fire and rescue missions that require large-scale monitoring and data collection. Multiple UAVs work together to provide more comprehensive perspectives and perception capabilities, enabling them to construct maps more quickly to support UGV navigation and task execution. In addition, multiple UAVs also increase the fault tolerance and robustness of the collaborative system to a certain extent. 

Stampa \emph{et al}. \cite{stampa2020scenario} designed an air-ground collaborative system consisting of multiple UAVs and a UGV as a mobile base station. 
In this system, the data collected by the UAVs are fused together, and then a 3-D map of the environment is constructed based on SLAM technology to assist the navigation of robots and human rescue missions. When the accurate environmental map is completed, the system switches to monitoring mode, with some UAVs providing a real-time aerial overview of the scene and quickly detecting environmental changes to effectively address the complexity and dynamics of the fire and rescue environment. Scaramuzza \emph{et al}. \cite{scaramuzza2014vision} described a collaborative system consisting of multiple UAVs, which can achieve autonomous navigation in large, unknown and GPS-free environments. These works mainly involve map construction of multi-UAV systems, but pay less attention to UGV navigation based on maps. For the path planning problem in multi-UAVs-UGV collaborative mode, Ren \emph{et al}. \cite{ren2018path} developed a framework for path planning in an air-ground collaborative system consisting of a pair of UAVs and a UGV. This architecture can be used for tasks such as collaborative search and rescue of heterogeneous robotic systems. Regarding the collaborative operation of UGV-UAV, Peng \emph{et al}. \cite{peng2019hybrid} proposed a routing and scheduling algorithm for multiple UAVs and a UGV, which allows multiple UAVs carried by the UGV to deliver multiple packages to different locations, providing useful insights for emergency supply delivery in fire and rescue missions. The above studies demonstrate the potential of multiple UAVs with a UGV for collaborative execution of fire and rescue missions in large-scale scenarios. Although multiple UAVs enhance environmental perception and rapid mapping, the single UGV may become a performance bottleneck in large-scale missions, particularly when task allocation efficiency, communication latency, and ground mobility constraints are jointly considered.


\subsection{Multi-UAVs with multi-UGVs}
Systems consisting of multiple UAVs and multiple UGVs are suitable for complex, large-scale fire and rescue missions, offering comprehensive environmental perception and efficient task execution. Multiple UAVs enable rapid deployment over fire-affected areas for search and mapping and can adapt their strategies to continue guiding UGV operations if individual UAVs fail. Multiple UGVs, in turn, facilitate simultaneous search and rescue operations at multiple locations and the delivery of rescue supplies.

Qin \emph{et al}. \cite{qin2019autonomous} designed a collaborative exploration and mapping method for UAVs and UGVs. The method uses UGVs to generate a rough environmental map based on a 2.5-D SLAM algorithm, and then the UAVs perform 3-D fine mapping to support task execution. 
Ghamry \emph{et al}. \cite{ghamry2016cooperative} developed a multi-UAV and multi-UGV collaborative system to monitor forests and detect fires. When a UAV detects a fire, it immediately transmits the relevant fire information to the leader UGV. After receiving the fire information, the leader UGV plans the reference trajectories for the UAV team based on the fire spread model, and the UAVs follow these trajectories to detect and continuously monitor the fire spread. 
Nazarova \emph{et al}. \cite{nazarova2020application} proposed applying the air-ground collaborative system to search and rescue missions to improve search and rescue efficiency. 
First, the sequence of rescue actions and rescue processes is analyzed based on statistical data. Then, based on probabilistic search theory and intelligent optimization algorithms, the air-ground collaborative system can complete tasks more efficiently. In order to achieve efficient collaboration between multiple UAVs and multiple UGVs, Phan \emph{et al}. \cite{phan2008cooperative} suggested a hierarchical architecture including a mobile mission controller, multiple UAVs and multiple UGVs, in which the UAVs are responsible for collecting environmental data and assisting in fire monitoring, while the UGVs are used to extinguish large open fires. Tanner \cite{tanner2007switched} combined the decentralized flocking method with the navigation function to develop an exchange cooperative control scheme to coordinate UAVs and UGVs to locate mobile targets in a given area. Through close collaboration among multiple UAVs and UGVs, mission efficiency and response capability can be significantly enhanced \cite{chai2024cooperative}. However, communication bandwidth management and distributed coordination pose increasing challenges as team size grows.

It is worth emphasizing that the four configurations discussed above are not merely variations in team size, but represent distinct organizational forms and collaboration mechanisms within a unified air-ground collaborative navigation paradigm. Specifically, a single UAV with a single UGV often reflects a leader-follower structure with unidirectional information flow. A single UAV with multi-UGVs extends this structure to one-to-many broadcast support. Multi-UAVs with a single UGV enable collaborative perception for a ground agent, and the multi-UAV-multi-UGV configuration can achieve fully distributed cooperative planning. As the team scale increases, the system evolves not only in size, but also in the coordination complexity and collaboration depth between aerial mapping and ground navigation.


\section{Application examples of air-ground collaborative robots} \label{section6}
Air-ground collaborative robots have demonstrated their advantages in firefighting and rescue missions, among which collaborative mapping and navigation technology plays an important role. This collaborative mode enables rapid response and efficient handling of complex scenarios such as forest wildfires, urban high-rise building fires, mountain rescue, and chemical plant fires, thereby improving the efficiency and safety of firefighting and rescue operations.

At present, air-ground collaborative robots have been implemented in practical applications and experimental tests in various scenarios. 
Frering \emph{et al}. \cite{frering2023multi} provided a collaborative robot system based on UAV and UGV for firefighting and rescue missions. 
The UAV uses RGB and thermal infrared cameras to detect hotspots and build a 3-D map of the fire scene. 
The UGV can extinguish the fire source with water based on the map information. 
The system is field-tested in the Seetaler Alps in Austria and demonstrates outstanding performance and reliability in terms of collaborative task completion, mission execution time, and robust operation under disturbances, validating the effectiveness of UAV-UGV cooperation in real-world firefighting scenarios.


To address forest fire and rescue tasks, Wei and Fang \cite{wei2024optimal} proposed a collaborative framework based on UAVs and UGVs. 
The UAVs use a coverage path planning algorithm to monitor the fire scene and guide the evacuation planning of UGVs, while the UGVs adopt a reactive algorithm to plan a collision-free path to collaboratively complete the evacuation task. Such lightweight planning is well suited to large-scale forest scenes requiring fast response, with a focus on computational efficiency. Results demonstrate effective task coordination, real-time obstacle avoidance, and scalability. 


For wildfire prevention, Pasini \emph{et al}. \cite{pasini2022uav} proposed a wildfire hotspot surveillance system based on UAV and UGV cooperation. The system uses the UAV to monitor and map the area after the fire from the air, and then uses the A* algorithm to plan a path for the UGV. 
The UGV conducts centimeter-level inspections along the planned path and employs a short-term correction algorithm to flexibly avoid unexpected obstacles. The system achieves accurate air-ground path transfer and tracking, and demonstrates stable navigation performance in dynamic environments.


For firefighting and rescue scenarios, Stampa \emph{et al}. \cite{stampa2020scenario} designed an air-ground collaborative system. 
In the system, the UAV uses RGB-D to build a 3-D map of the environment and detect fires. The UGV performs path planning and navigation based on the map built by the UAV and drags a hose to extinguish the fire. 
Similarly, Martinez-Rozas \emph{et al}. \cite{martinez2022aerial} introduced a UAV/UGV collaborative team for autonomous firefighting in urban environments, consisting of three UAVs and one UGV. The UAVs take advantage of 3-D LiDAR for mapping and use infrared thermal imagers to sense the location of the fire. The UGV utilizes a combination of global path planning and local path tracking for navigation. This system was verified in the high-rise fire rescue scenarios, demonstrating reliable navigation and effective task execution. 


\begin{table*}[!t]
	\renewcommand{\arraystretch}{1.2}
	\setlength{\tabcolsep}{2pt}
	\centering
	\caption{Comparison of representative air-ground collaborative systems in fire and rescue applications}
	\label{table4}
	\scriptsize
	\begin{tabular*}{\textwidth}{
			m{1.5cm}<{\centering}
			m{2.3cm}<{\centering}
			m{1.8cm}<{\centering}
			m{1.0cm}<{\centering}
			m{2.5cm}<{\centering}
			m{3.6cm}<{\centering}
			m{4.5cm}<{\centering}
		}
		\toprule
		Reference &
		Scenario &
		Team &
		Map Type &
		Navigation Strategy &
		Key Indicators &
		Failure Modes \\
		\midrule
		
		Frering \emph{et al}. \cite{frering2023multi} 
		& Wildfire, mountain terrain
		& 1 UAV + 1 UGV
		& 3-D 
		& Autonomous navigation
		& Mission duration: 28--41 min; Attention Demand: 29--52\%
		& Hotspot misdetection; Communication dropouts \\
		\cmidrule(lr){1-7} 
		Wei and Fang \cite{wei2024optimal}
		& Forest fire simulation
		& Multiple UAVs + UGVs
		& 2-D
		& Coverage planning + reactive navigation
		& Fire detection: 47.5 s; first rescue: 52.25 s; second rescue: 68.5 s; safe point: 85.1 s
		& Limited real-world validation \\
		\cmidrule(lr){1-7} 
		Pasini \emph{et al}. \cite{pasini2022uav}
		& Post-fire hotspot inspection
		& 1 UAV + 1 UGV
		& 2-D
		& A* + short-term correction
		& Xtrack error: mean $<$0.003 m, std $<$0.048 m
		& GPS drift; Terrain misclassification under low contrast \\
		\cmidrule(lr){1-7} 
		Stampa \emph{et al}. \cite{stampa2020scenario}
		& Industrial complex fire
		& Multiple UAVs + mobile GCS
		& 3-D
		& SLAM-based exploration (conceptual)
		& - 
		& Potential communication loss; Sensor disturbances under heat and smoke \\
		\cmidrule(lr){1-7} 
		Martinez-Rozas \emph{et al}. \cite{martinez2022aerial}
		& Urban high-rise fire
		& 3 UAVs + 1 UGV
		& 3-D
		& Lazy Theta* + local path tracking
		& Localization RMSE: 0.217 m; local planning rate: 7 Hz; MBZIRC scores: 5.0, 3.5, and 0.13
		& GPS signal loss; Wind gusts; Water-aiming errors; Magnet release failure \\
		
		\bottomrule
	\end{tabular*}
\end{table*}
For ease of comparison, representative examples are summarized in Table \ref{table4}.
In summary, air-ground collaborative robots have achieved remarkable results in fire and rescue. They use UAVs for environmental perception and mapping to support UGV navigation and task execution, thereby improving rescue efficiency and reducing risks to operators.

\section{Conclusion and Outlook} \label{section7}
The mapping and navigation technology of air-ground collaborative robots is the core foundation for realizing the efficient and autonomous collaboration of UAVs and UGVs to complete fire and rescue tasks. To this end, this paper provides a systematic review of mapping and navigation technologies for air-ground collaborative robots in fire and rescue missions from a new perspective. Centered on a collaborative framework where UAVs construct environment maps and UGVs perform path planning and navigation based on these maps, we summarize representative approaches from three key aspects: UAV-based map construction
, collaborative localization methods, and UGV navigation and path planning strategies under different map representations. Furthermore, typical air-ground collaboration configurations with different numbers of UAVs and UGVs are reviewed, and their respective advantages and limitations in fire and rescue scenarios are analyzed. Representative application cases are also discussed to demonstrate the practical value of air-ground collaborative robots in real-world fire and rescue tasks. Overall, this paper can provide a valuable reference for the deployment and application of air-ground collaborative robots to improve rescue efficiency and personnel safety in fire and rescue missions.

Although air-ground collaborative robots have shown promising potential in fire and rescue missions, significant challenges remain in enabling UAVs to construct task-relevant maps under highly dynamic and degraded conditions, as well as ensuring robust UGV navigation and task execution in complex and uncertain disaster environments. For fire and rescue missions, the following research directions represent promising opportunities for advancing air-ground collaborative robot mapping and navigation.

1) Fire and rescue mission-oriented UAV mapping. 
Existing maps mostly focus on the map itself, ignoring the relevance to the fire and rescue missions. Establishing an environment model that highlights task-related environmental features, semantics and other information has been shown to significantly enhance robotic task execution efficiency \cite{zhang2024Environment, wu2023object}. 
Thus, integrating the operating environment and task characteristics, leveraging technologies such as task point detection, semantic segmentation, heat map analysis, and 3-D reconstruction 
to construct maps that facilitate UGV navigation and task execution is a promising research direction. Such maps can provide critical information, including fire distribution, hotspot locations, and building structures, to support rescue planning.

2) Multi-modal data fusion for UAV mapping. Fire and rescue environments are often complex and highly dynamic, featuring smoke, flames, and obstacles, which pose significant challenges to UAV mapping. Multi-sensor SLAM systems have demonstrated strong robustness and accuracy in such complex settings \cite{du2025advancements}, 
and can effectively mitigate the adverse effects of smoke and illumination variations on visual perception \cite{marques2024applying}. To this end, heterogeneous sensor modalities, such as thermal, visible-light, and infrared data, can be integrated to support UAV mapping. Such multi-modal fusion enhances the robustness and completeness of UAV perception and mapping, while providing richer environmental representations to facilitate UGV navigation. Consequently, UAV mapping systems based on multi-modal sensor fusion are expected to play an increasingly important role in air-ground collaborative robots for fire and rescue missions.

3) UGV navigation based on UAV maps. Based on different types of maps built by UAVs, it is particularly important to study the general navigation of UGVs, which is the basis for realizing the execution of air-ground collaborative robot tasks. With the continuous integration of deep learning technology, the localization and perception capabilities of the UGV in unknown or unstructured environments are enhanced \cite{soori2023artificial, zhang2026human}. On this basis, by learning environmental features in real time, UGVs can achieve fast path planning and efficient obstacle avoidance based on different types of maps, further enhancing the adaptability of UGV navigation. Note that the improvement of edge computing capabilities can effectively improve the efficiency and real-time performance of UGV navigation based on deep learning, which will be a direction worth exploring for air-ground collaborative robots.

4) Embodied artificial intelligence (AI)-driven UGV navigation. The integration of embodied AI and robotics has significantly impacted robot navigation \cite{liu2024aligning}. Research has shown that incorporating embodied AI, particularly large language models (LLMs) and visual language models (VLMs), substantially improves robots' scene understanding, path planning, and autonomous navigation in complex environments \cite{zhang2025zisvfm, wang2026survey}. 
In fire and rescue missions, the combination of collaborative navigation systems and embodied AI can, on the one hand, improve the performance of the UAV in real-time fire scene understanding, task area identification and feature matching for UGV navigation, and on the other hand, effectively enhance the navigation performance and mission execution capabilities of the UGV. This will play an integral role in air-ground collaborative robot navigation.

\bibliographystyle{IEEEtran}
\bibliography{mybibfile}

\vspace{-3em}
\begin{IEEEbiography}[{\includegraphics[width=1in,height=1.25in,clip,keepaspectratio]{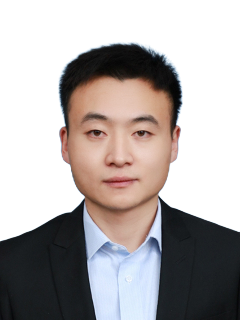}}]{Ying Zhang} (Senior Member, IEEE) received the Ph.D. degree in Control Theory and Control Engineering from Shandong University, Jinan, China, in 2021.

He is currently an Associate Professor with the Yanshan University. He is an Associate Editor / Early Career Associate Editor for \emph{International Journal of Advanced Robotic Systems}, \emph{Biomimetic Intelligence and Robotics}, \emph{Intelligence \& Robotics}, \emph{SmartBot}, etc. Dr. Zhang received Best Paper Award in CCIR 2018 and ICBIR 2026. His current research interests include intelligent robot systems, environment modeling, embodied perception and robot skill learning.
\end{IEEEbiography}

\vspace{-3em}
\begin{IEEEbiography}[{\includegraphics[width=1in,height=1.25in,clip,keepaspectratio]{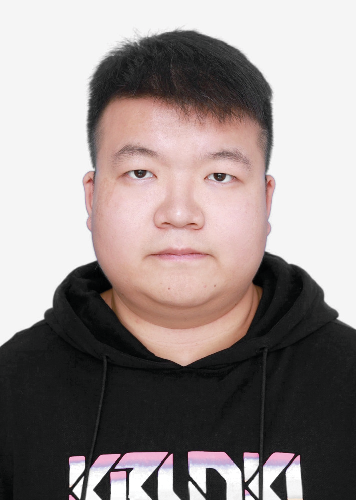}}]{Haibao Yan} received the B.S. degree in automation from Taiyuan University of Technology, Taiyuan, China, in 2021. He is currently pursuing the M.S. degree in control engineering at Yanshan University, Qinhuangdao, China. 

His current research interests include air-ground collaborative robots and simultaneous localization and mapping (SLAM).
\end{IEEEbiography}

\vspace{-3em}
\begin{IEEEbiography}[{\includegraphics[width=1in,height=1.25in,clip,keepaspectratio]{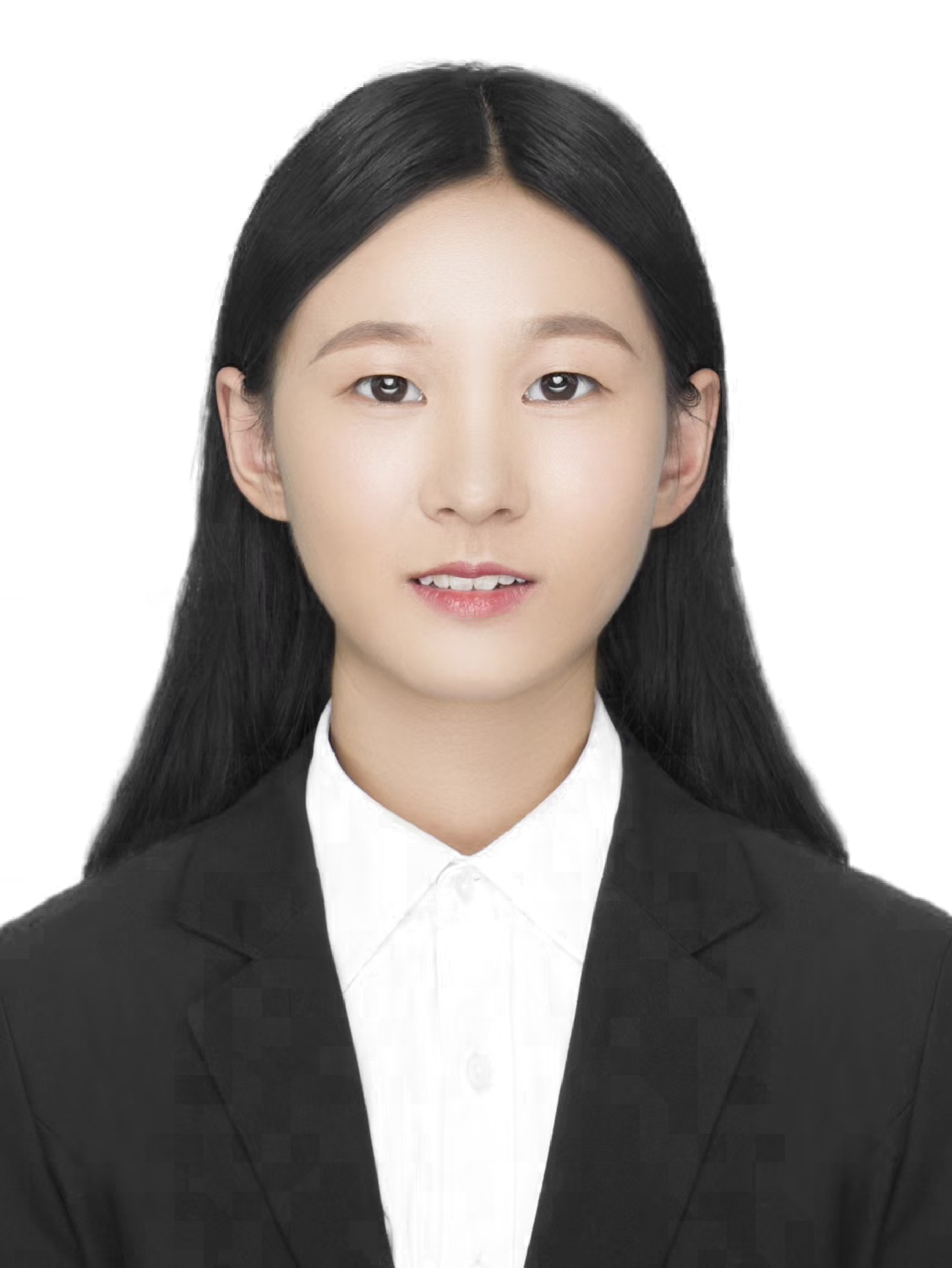}}]{Danni Zhu} received the B.S. degree in automation from Yanshan University in 2021. She is currently pursuing the M.S. degree in control engineering at Yanshan University, Qinhuangdao, China. 

Her current research interests include air-ground cooperative robots and path planning.
\end{IEEEbiography}

\vspace{-3em}
\begin{IEEEbiography}[{\includegraphics[width=1in,height=1.25in,clip,keepaspectratio]{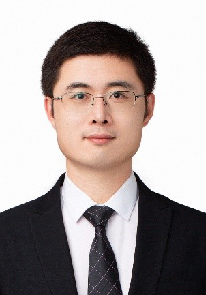}}]{Jiankun Wang} (Senior Member, IEEE) received the B.E. degree in automation from Shandong University, Jinan, China, in 2015, and the Ph.D. degree from the Department of Electronic Engineering, The Chinese University of Hong Kong, Hong Kong, in 2019. During the Ph.D. degree, he spent six months with Stanford University, Stanford, CA, USA, as a Visiting Student Scholar, supervised by Prof. Oussama Khatib. 

He is currently an Assistant Professor with the Department of Electronic and Electrical Engineering, Southern University of Science and Technology, Shenzhen, China. He is an Associate Editor for the \textsc{IEEE Transactions on Automation Science and Engineering}, \textsc{IEEE Transactions on Intelligent Vehicles}, \textsc{IEEE Robotics and Automation Letters}, \emph{International Journal of Robotics and Automation}, and \emph{Biomimetic Intelligence and Robotics}. His current research interests include motion planning and control, human-robot interaction, and machine learning in robotics.
\end{IEEEbiography}

\vspace{-3em}
\begin{IEEEbiography}[{\includegraphics[width=1in,height=1.25in,clip,keepaspectratio]{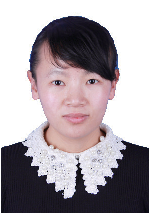}}]{Cui-Hua Zhang} (Member, IEEE) received the M.S. degree in control theory and control engineering from the Institute of Automation, Qufu Normal University, Qufu, China, in 2017, and the Ph.D. degree in control theory and control engineering from Northeastern University, Shenyang, China, in 2021. 

She is currently an Associate Professor with the Yanshan University. He is an Early Career Associate Editor for \emph{Journal of Automation and Intelligence}. Her current research interests include intelligent robot systems, and networked nonlinear systems.
\end{IEEEbiography}

\vspace{-3em}
\begin{IEEEbiography}[{\includegraphics[width=1in,height=1.25in,clip,keepaspectratio]{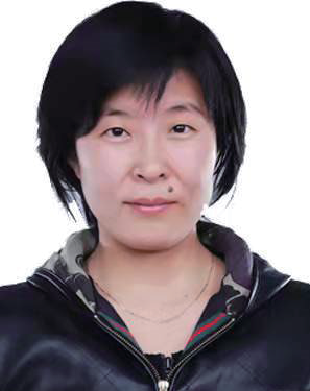}}]{Weili Ding} received the B.Sc. degree in automation from the Liaoning University of Technology, Jinzhou, China, in 2003, and the Ph.D. degree from the Shenyang Institute of Automation, Chinese Academy of Sciences, Beijing, China, in 2008.

She is currently a Full Professor with the Department of Automation, Yanshan University, Qinhuangdao, China. Her research interests include machine vision, image processing, and virtual reality.
\end{IEEEbiography}
\vspace{-3em}
\begin{IEEEbiography}[{\includegraphics[width=1in,height=1.25in,clip,keepaspectratio]{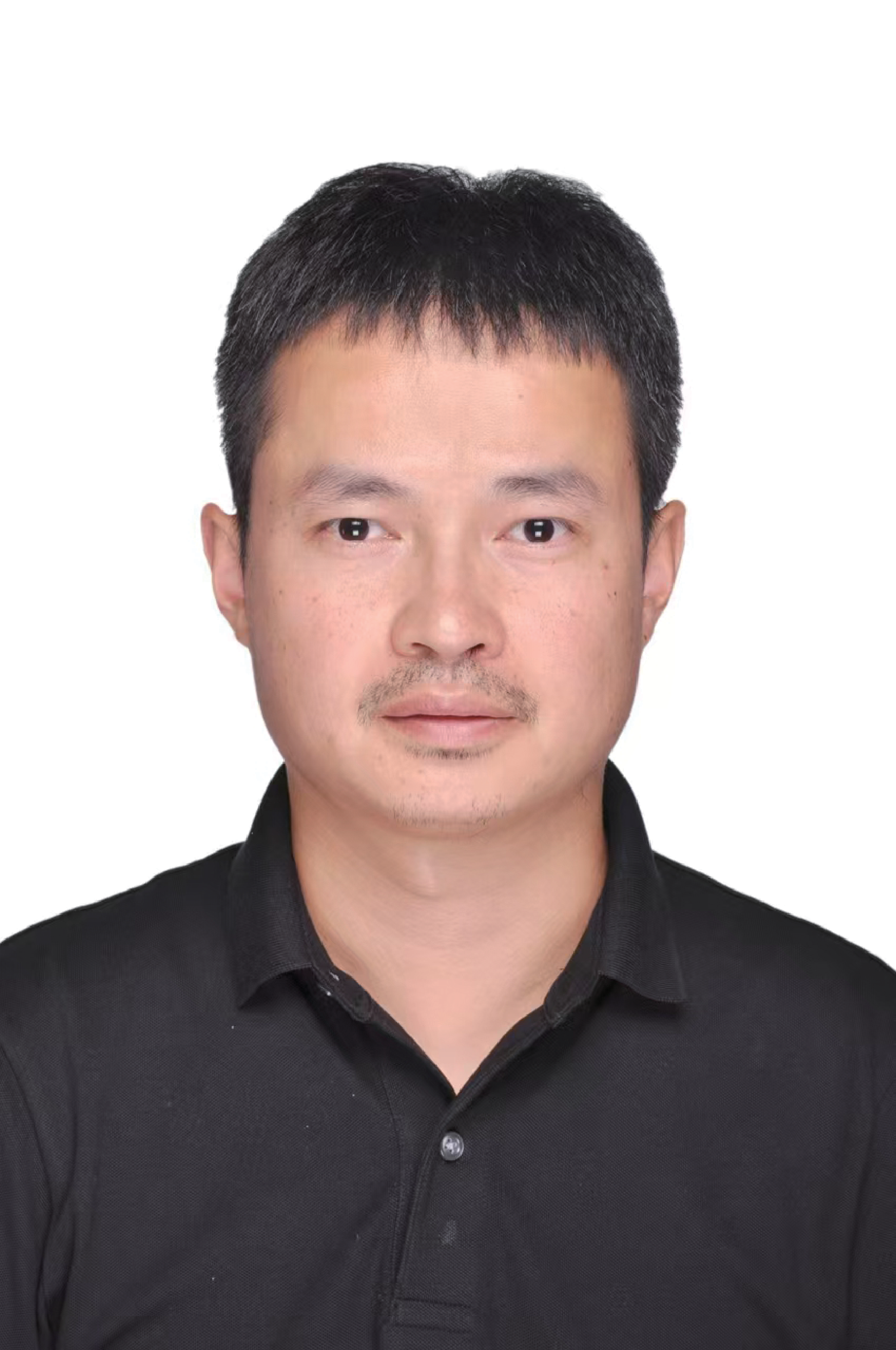}}]{Xi Luo} (Member, IEEE)
received the M.S. and Ph.D. degrees in electrical engineering from Yanshan University, Qinhuangdao, China, in 2014 and 2023, respectively. 

He is currently an Engineer with the School of Electrical Engineering, Yanshan University. His research interests include nonlinear control, tracking control of unmanned aerial vehicles and underwater vehicles.
\end{IEEEbiography}

\vspace{-3em}
\begin{IEEEbiography}[{\includegraphics[width=1in,height=1.25in,clip,keepaspectratio]{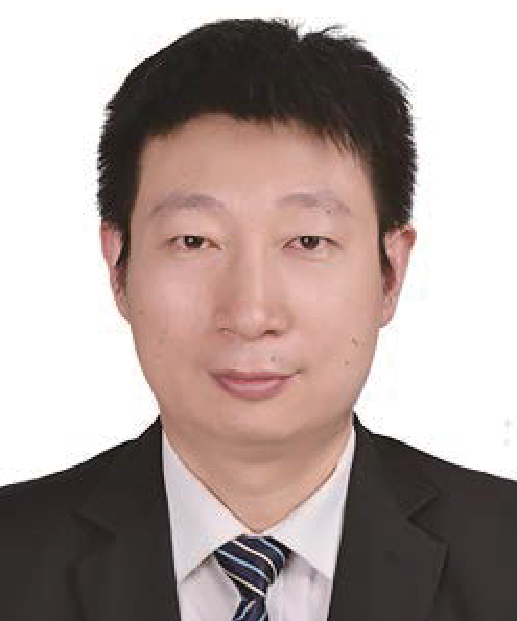}}]{Changchun Hua} (Fellow, IEEE) received the Ph.D. degree in Electrical Engineering from Yanshan University, Qinhuangdao, China, in 2005. From 2006 to 2007, he was a Research Fellow with the National University of Singapore, Singapore. From 2007 to 2009, he was with Carleton University, Ottawa, ON, Canada, funded by the Province of Ontario Ministry of Research and Innovation Program. From 2009 to 2010, he was with the University of Duisburg-Essen, Essen, Germany, funded by the Alexander von Humboldt Foundation.

He is currently a Full Professor with Yanshan University. He is a Cheung Kong Scholars Programme Special Appointment Professor. He is an Associate Editor for the \textsc{IEEE Transactions on Cybernetics}, \emph{International Journal of Control, Automation and Systems}, and other journals. He is a highly cited Researcher (since 2014) selected by Elsevier. His research interests include nonlinear control systems, multiagent systems, teleoperation systems, and intelligent robot systems.
\end{IEEEbiography}
\vspace{-3em}
\begin{IEEEbiography}[{\includegraphics[width=1in,height=1.25in,clip,keepaspectratio]{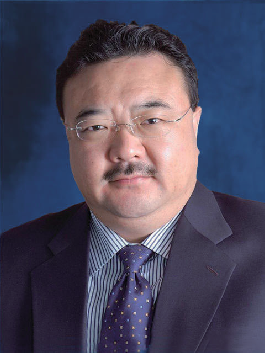}}]{Max Q.-H. Meng} (Fellow, IEEE) received the Ph.D. degree in electrical and computer engineering from the University of Victoria, Canada, in 1992. He joined The Chinese University of Hong Kong, as a Professor, in 2001, and later the Chairman of the Department of Electronic Engineering. He was with the Department of Electrical and Computer Engineering, University of Alberta, Canada, where he was the Director of the Advanced Robotics and Teleoperation (ART) Laboratory and held the positions of an Assistant Professor in 1994, an Associate Professor in 1998, and a Professor in 2000. He was an Academician of the Canadian Academy of Engineering. 

He is currently a Chair Professor and the Head of the Department of Electronic and Electrical Engineering, Southern University of Science and Technology, Shenzhen, China. He is also an Honorary Chair Professor with the Harbin Institute of Technology and Zhejiang University, and also the Honorary Dean of the School of Control Science and Engineering, Shandong University, China. He has published more than 750 journal articles and conference papers and book chapters and led more than 60 funded research projects to completion as a principal investigator. His research interests include medical and service robotics, and robotics perception and intelligence. He served as an elected member for the AdCom of IEEE RAS for two terms. He was a Fellow of the Hong Kong Institution of Engineers. He was a recipient of the IEEE Millennium Medal. He was the General Chair or the Program Chair of many international conferences, including the General Chair of IROS 2005 and ICRA 2021. He served as the Co-Chair for the Fellow Evaluation Committee. He served as an Associate VP for Conferences of the IEEE Robotics and Automation Society from 2004 to 2007. He has been serving as the Editor-in-Chief and editorial board for a number of international journals, including the Editor-in-Chief for the \emph{Biomimetic Intelligence and Robotics} (Elsevier).
\end{IEEEbiography}

\end{document}